\documentclass{article}
\usepackage{spconf,amsmath,amssymb,graphicx,bm}
\usepackage{url}
\usepackage{enumitem}
\setlist{nosep, leftmargin=12pt}
\usepackage{booktabs}
\usepackage[table]{xcolor}
\usepackage{pifont}
\usepackage{multirow}
\usepackage{algorithm}
\usepackage{algpseudocode}
\definecolor{oursbg}{RGB}{238,238,238}
\newcommand{\vv}[1]{\mathbf{#1}}

\makeatletter\@secpenalty=0\makeatother

\title{Dual-VaDE: Efficient and Robust Prototype Learning via Variational Dual-Domain Mixture Priors}

\name{Wugeng Zheng$^{1}$, Song Wang$^{3}$, Jian Wang$^{2,3}$}
\address{$^{1}$College of Engineering, Northeastern University, Boston, MA, USA\\
$^{2}$Athinoula A. Martinos Center for Biomedical Imaging, Department of Radiology,  \\Massachusetts General Hospital, Harvard Medical School, Boston, MA, USA \\
$^{3}$Computer Science, University of Central Florida, Orlando, FL, USA}

\begin{document}
\maketitle

\begin{abstract}
In this paper, we propose Dual-VaDE, an unsupervised generative framework for learning interpretable anatomical prototypes from unlabeled medical images. In contrast to previous variational methods that rely on a single spatial latent code, which often entangles global geometry with local textures and fails to capture their cross-domain dependencies, Dual-VaDE introduces a dual-domain Gaussian mixture prior that explicitly couples spatial anatomy with phase-preserving Fourier spectral representations. We derive an exact, closed-form variational lower bound that fuses this joint prior into the decoder and optimizes the entire network end-to-end via an alternating EM formulation, avoiding heuristic two-stage clustering while ensuring superior training stability and robustness. Evaluated on OASIS-3 brain MRI, Dual-VaDE delivers substantially sharper and more plausible prototypes, achieves superior cluster separation with improvements of up to 94\% over spatial-only baselines, exhibits markedly reduced variance across random seeds, and incurs negligible inference overhead (0.28~ms per image).\footnote{Source code is available at \url{https://github.com/ZhengWugeng/Dual-VaDE}; the package can be installed via \texttt{pip install dual-vade}.}
\end{abstract}


\section{Introduction}
\label{sec:introduction}

High-dimensional medical imaging exhibits complex morphological variability that conventional unsupervised representation learning often fails to disentangle. Unlike supervised pipelines built on predefined categories or diagnostic labels, unsupervised distribution modeling can capture the continuous spectrum of anatomical heterogeneity directly from raw cohorts~\cite{habes2020heterogeneity}. Deep generative models enable density estimation and unsupervised representation learning without manual annotations~\cite{kingma2014vae,ho2020ddpm,kazerouni2023diffusion, evan2018convolutional}. Compared with discriminative clustering, generative methods can improve interpretability by decoding latent representations into anatomical prototypes—inspectable templates that support subsequent morphological analysis~\cite{yang2021smilegan}.

Variational autoencoders (VAEs) with Gaussian mixture priors~\cite{jiang2017vade,dilokthanakul2016gmvae,falck2021multi} link each mixture component to a discrete morphological centroid, where decoding the mean yields a prototype. However, typical formulations use a single spatial latent vector, entangling global geometry (e.g., compartment volume) with local appearance (e.g., contrast and boundary texture). With unstructured diagonal or unimodal covariances, such priors fail to capture cross-scale geometry–texture dependencies, often producing blurred or degenerate prototypes. The Fourier domain offers a complementary view of the image manifold: low frequencies largely determine spectral energy, while phase preserves topology, boundaries, and edges~\cite{oppenheim1981phase}. Magnitude-only models drop this geometric phase information, whereas complex-valued representations~\cite{trabelsi2018complex,stanic2023complex,gopalakrishnan2024complex} or real–imaginary decompositions~\cite{chi2020ffc,rao2021gfnet} retain full anatomical content. Although dual-domain fusion is popular~\cite{frank2020leveraging,rippel2015spectral,xu2020learning}, deterministic frequency features alone are insufficient and often redundant with spatial cues (Sec.~\ref{sec:quantitative}). Learning coherent morphological prototypes requires a structured prior that explicitly couples spatial and spectral latents.

We present Dual-VaDE, an unsupervised prototype-learning framework that factorizes latents into coupled spatial and spectral subspaces. Our contributions are:
\begin{enumerate}
\item \textbf{Dual-domain mixture prior:} We propose a dual-domain Gaussian mixture prior, parameterized by a newly designed mean and covariance matrix whose off-diagonal cross-covariance terms explicitly capture within-prototype spatial–spectral co-variation.
\item \textbf{End-to-End Variational EM Optimization:} We derive a new closed-form variational lower bound, optimized end-to-end via EM, avoiding multi-stage error accumulation and improving training stability and clustering robustness.
\item \textbf{Empirical evaluation and open-source implementation:} On structural brain MRI, Dual-VaDE yields sharper prototypes, better cluster separability, and more stable optimization than eight generative baselines with minimal overhead.
\end{enumerate}

\begin{figure*}[!t]
\centering
\includegraphics[width=\textwidth]{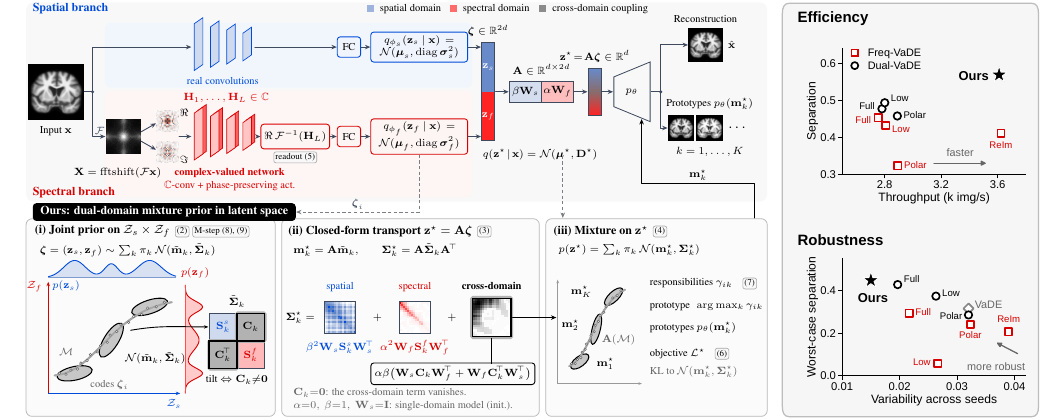}
\caption{Dual-VaDE overview. Left: a spatial encoder and a phase-preserving spectral encoder (real/imaginary Fourier channels) produce two latent codes. A dual-domain mixture prior couples them via the cross-domain block $\vv C_k$. The fused code feeds a shared decoder to generate reconstructions and prototypes. Right: prototype separation (mean Silhouette over $K$) vs. throughput, and robustness measured as seed-to-seed variability under the worst case over $K$.}
\label{fig:model}
\end{figure*}

\section{Background}
\label{sec:background}

Let $\vv x$ be a medical image, $\vv z\in\mathbb{R}^d$ a continuous latent code, and $c\in\{1,\dots,K\}$ a discrete prototype index. VaDE~\cite{jiang2017vade} assumes the generative process $c\sim\mathrm{Cat}(\bm\pi)$, $\vv z\mid c{=}k\sim\mathcal{N}(\vv m_k,\bm\Sigma_k)$, and $\vv x\mid\vv z\sim p_\theta(\vv x\mid\vv z)$, with an amortized posterior $q_\phi(\vv z\mid\vv x)=\mathcal{N}(\bm\mu,\vv D)$ and $\vv D=\mathrm{diag}(\bm\sigma^2_\phi(\vv x))$. The evidence lower bound (ELBO) is
\begin{equation}
\begin{aligned}
\mathcal{L}=\;&\mathbb{E}_{q_\phi}[\log p_\theta(\vv x\mid \vv z)]-\mathrm{KL}(\bm\gamma\,\|\,\bm\pi)\\
&-\textstyle\sum_{k=1}^{K}\gamma_k\,\mathrm{KL}\big(\mathcal{N}(\bm\mu,\vv D)\,\|\,\mathcal{N}(\vv m_k,\bm\Sigma_k)\big),
\end{aligned}
\label{eq:standard_elbo}
\end{equation}
where $\bm\gamma$ are the posterior prototype responsibilities~\cite{jiang2017vade,dilokthanakul2016gmvae,xie2016unsupervised}, and related unsupervised clustering methods are explored in~\cite{xie2016unsupervised}. Because a single code $\vv z$ must capture both global anatomy and local texture, each $\bm\Sigma_k$ is an unstructured $d\times d$ matrix without factor blocks, and the model cannot represent cross-domain dependencies explicitly. Multi-view generative models address a related problem by assigning one latent code per view~\cite{shi2019mmvae}; we follow this idea within the prior by giving it two codes whose covariance blocks follow this split.
\vspace{-4.2mm}
\section{Methodology: Dual-VaDE}
\label{sec:method}

\subsection{Dual-domain mixture prior}
\label{sec:cross-domain-covariance}

We keep the generative process of Sec.~\ref{sec:background} and change only the prior. Each image is embedded in a joint spatial and spectral latent space by concatenating a spatial code $\vv z_s$ and a spectral code $\vv z_f$, both in $\mathbb{R}^{d}$, into $\bm\zeta=[\vv z_s;\vv z_f]\in\mathbb{R}^{2d}$, on which we impose a Gaussian mixture with block-structured parameters:
\begin{equation}
\begin{aligned}
\bm\zeta\mid c{=}k &\sim \mathcal{N}(\tilde{\vv m}_k,\tilde{\bm\Sigma}_k),\\
\tilde{\vv m}_k &= \begin{bmatrix} \vv m_k^{s} \\ \vv m_k^{f} \end{bmatrix},\quad
\tilde{\bm\Sigma}_k = \begin{bmatrix} \vv S_k^{s} & \vv C_k \\ \vv C_k^{\top} & \vv S_k^{f} \end{bmatrix}.
\end{aligned}
\label{eq:hybrid}
\end{equation}
Here, $\vv S_k^{s}$ and $\vv S_k^{f}$ are the within-prototype spatial and spectral covariances, and the cross-domain block $\vv C_k\in\mathbb{R}^{d\times d}$ captures how anatomy and spectral content co-vary within prototype $k$.

To obtain a single generative path, we define $\vv z^\star=\vv A\bm\zeta$ with $\vv A=[\,\beta \vv W_s\;\;\alpha \vv W_f\,]$, learnable $\vv W_s,\vv W_f\in\mathbb{R}^{d\times d}$, and scalars $\alpha,\beta$. A linear map of a Gaussian mixture is again a Gaussian mixture with unchanged weights~\cite{bishop2006prml}, whose parameters are
\begin{equation}
\begin{aligned}
\vv m_k^\star =\;& \beta \vv W_s \vv m_k^{s} + \alpha \vv W_f \vv m_k^{f}, \\
\bm\Sigma_k^\star =\;& \beta^2 \vv W_s \vv S_k^{s} \vv W_s^{\top} + \alpha^2 \vv W_f \vv S_k^{f} \vv W_f^{\top} \\
&+ \alpha\beta\big(\vv W_s \vv C_k \vv W_f^{\top} + \vv W_f \vv C_k^{\top} \vv W_s^{\top}\big).
\end{aligned}
\label{eq:induced}
\end{equation}
The covariance thus decomposes exactly into a spatial, a spectral, and a cross-domain term, the last of which vanishes when $\vv C_k=\vv 0$. The resulting \emph{dual-domain mixture prior} on the decoder input is
\begin{equation}
p(\vv z^\star) = \textstyle\sum_{k=1}^K \pi_k \, \mathcal{N}\big(\vv z^\star \mid \vv m_k^\star, \bm\Sigma_k^\star \big),
\label{eq:new_gaussian_prior}
\end{equation}
which reduces to the VaDE prior for $\alpha=0$, $\beta=1$, and $\vv W_s=\vv I$; we use this setting as initialization.

\subsection{Phase-preserving spectral encoder}
\label{sec:phase-preserving-reim}

The spectral code $\vv z_f$ is computed from the centered spectrum $\vv X=\mathrm{fftshift}(\mathcal{F}(\vv x))\in\mathbb{C}^{H\times W}$. Since magnitude-only inputs $g(|\vv X|)$ are invariant to phase and thus discard spatial localization~\cite{oppenheim1981phase}, we represent $\vv X$ by its real and imaginary parts as two channels, as in recent spectral networks~\cite{chi2020ffc,rao2021gfnet}, on which the complex-valued layers~\cite{trabelsi2018complex} are implemented\footnote{A complex convolution $(a+ib)*(x+iy)$ equals the real convolution of $[x;y]$ with the weights $\left[\begin{smallmatrix}a&-b\\b&a\end{smallmatrix}\right]$, so real-valued weights on the two channels represent it exactly and are strictly more general.}. The final complex feature map $\vv H_L$ is converted to real features by the real-part readout
\begin{equation}
\begin{aligned}
\mathcal{R}(\vv H_L) &= \Re\big(\mathcal{F}^{-1}(\vv H_L)\big) = \mathcal{F}^{-1}(\vv H_L^{\mathrm{h}}), \\
\vv H_L^{\mathrm{h}}(\omega) &= \tfrac{1}{2}\big(\vv H_L(\omega) + \overline{\vv H_L(-\omega)}\big),
\end{aligned}
\label{eq:readout}
\end{equation}
which is linear and injective on Hermitian-symmetric spectra and therefore preserves both the magnitude and the relative phase of $\vv H_L^{\mathrm{h}}$.

\begin{table*}[t]
\centering
\caption{Prototype separation on OASIS-3 (mean$\pm$SD, 3 seeds). \emph{Silhouette} ($\uparrow$, $[-1,1]$): closer to own than nearest other prototype. \emph{Davies-Bouldin} ($\downarrow$): within-prototype scatter vs between-prototype distance. Phase: Fourier phase preserved. Dual-VaDE augments the cited spectral encoder with a spatial branch and our dual-domain prior. \textbf{Bold}/\underline{underline}: best/second best. $^{\dagger}p<0.05$, $^{\ddagger}p<0.01$: ours significantly better (one-sided Welch’s $t$-test).}
\label{tab:oasis-full}
\footnotesize
\setlength{\tabcolsep}{2.4pt}
\renewcommand{\arraystretch}{1.1}
\begin{tabular}{@{}llccccccc c@{}}
\toprule
& & & \multicolumn{3}{c}{Silhouette $\uparrow$} & \multicolumn{3}{c}{Davies-Bouldin $\downarrow$} & \\
\cmidrule(lr){4-6}\cmidrule(lr){7-9}
Model & Spectral input & Phase & $K=3$ & $K=5$ & $K=7$ & $K=3$ & $K=5$ & $K=7$ & Avg.\ rank \\
\midrule
VaDE~\cite{jiang2017vade} & None &  & $0.466${\scriptsize$\,\pm$0.022}$^{\ddagger}$ & $0.315${\scriptsize$\,\pm$0.009}$^{\ddagger}$ & $0.393${\scriptsize$\,\pm$0.065}$^{\phantom{\ddagger}}$ & $0.841${\scriptsize$\,\pm$0.083}$^{\ddagger}$ & $1.179${\scriptsize$\,\pm$0.252}$^{\dagger}$ & $0.717${\scriptsize$\,\pm$0.196}$^{\phantom{\ddagger}}$ & 6.67 \\
\midrule
\multirow{4}{*}{Freq-VaDE} & FullFreq~\cite{frank2020leveraging} & \ding{55} & $0.594${\scriptsize$\,\pm$0.003}$^{\dagger}$ & $0.470${\scriptsize$\,\pm$0.023}$^{\ddagger}$ & $0.293${\scriptsize$\,\pm$0.039}$^{\ddagger}$ & $0.493${\scriptsize$\,\pm$0.006}$^{\phantom{\ddagger}}$ & $0.636${\scriptsize$\,\pm$0.024}$^{\ddagger}$ & $0.897${\scriptsize$\,\pm$0.101}$^{\dagger}$ & 4.33 \\
 & LowFreq~\cite{rippel2015spectral} & \ding{55} & $\underline{0.637}${\scriptsize$\,\pm$0.021}$^{\phantom{\ddagger}}$ & $\underline{0.600}${\scriptsize$\,\pm$0.013}$^{\phantom{\ddagger}}$ & $0.058${\scriptsize$\,\pm$0.046}$^{\ddagger}$ & $\underline{0.480}${\scriptsize$\,\pm$0.022}$^{\phantom{\ddagger}}$ & $\mathbf{0.449}${\scriptsize$\,\pm$0.011}$^{\phantom{\ddagger}}$ & $2.003${\scriptsize$\,\pm$0.348}$^{\ddagger}$ & 4.25 \\
 & PolarFreq~\cite{gopalakrishnan2024complex} & \ding{51} & $0.415${\scriptsize$\,\pm$0.006}$^{\ddagger}$ & $0.314${\scriptsize$\,\pm$0.036}$^{\ddagger}$ & $0.240${\scriptsize$\,\pm$0.055}$^{\ddagger}$ & $1.111${\scriptsize$\,\pm$0.040}$^{\ddagger}$ & $1.168${\scriptsize$\,\pm$0.034}$^{\ddagger}$ & $1.106${\scriptsize$\,\pm$0.086}$^{\ddagger}$ & 8.33 \\
 & ReIm~\cite{stanic2023complex} & \ding{51} & $0.589${\scriptsize$\,\pm$0.036}$^{\dagger}$ & $0.438${\scriptsize$\,\pm$0.032}$^{\ddagger}$ & $0.206${\scriptsize$\,\pm$0.049}$^{\ddagger}$ & $0.546${\scriptsize$\,\pm$0.029}$^{\ddagger}$ & $0.685${\scriptsize$\,\pm$0.029}$^{\ddagger}$ & $1.082${\scriptsize$\,\pm$0.310}$^{\dagger}$ & 6.17 \\
\midrule
\multirow{3}{*}{Dual-VaDE} & FullFreq~\cite{frank2020leveraging} & \ding{55} & $0.550${\scriptsize$\,\pm$0.009}$^{\ddagger}$ & $0.453${\scriptsize$\,\pm$0.044}$^{\ddagger}$ & $\underline{0.427}${\scriptsize$\,\pm$0.006}$^{\dagger}$ & $0.596${\scriptsize$\,\pm$0.013}$^{\ddagger}$ & $0.691${\scriptsize$\,\pm$0.085}$^{\dagger}$ & $\underline{0.703}${\scriptsize$\,\pm$0.070}$^{\dagger}$ & 5.25 \\
 & LowFreq~\cite{rippel2015spectral} & \ding{55} & $0.583${\scriptsize$\,\pm$0.013}$^{\ddagger}$ & $0.523${\scriptsize$\,\pm$0.011}$^{\ddagger}$ & $0.373${\scriptsize$\,\pm$0.055}$^{\phantom{\ddagger}}$ & $0.574${\scriptsize$\,\pm$0.088}$^{\phantom{\ddagger}}$ & $0.599${\scriptsize$\,\pm$0.032}$^{\ddagger}$ & $0.703${\scriptsize$\,\pm$0.098}$^{\phantom{\ddagger}}$ & \underline{4.08} \\
 & PolarFreq~\cite{gopalakrishnan2024complex} & \ding{51} & $0.576${\scriptsize$\,\pm$0.010}$^{\ddagger}$ & $0.511${\scriptsize$\,\pm$0.051}$^{\dagger}$ & $0.285${\scriptsize$\,\pm$0.035}$^{\ddagger}$ & $0.538${\scriptsize$\,\pm$0.012}$^{\dagger}$ & $0.563${\scriptsize$\,\pm$0.080}$^{\phantom{\ddagger}}$ & $0.959${\scriptsize$\,\pm$0.091}$^{\ddagger}$ & 4.83 \\
\textbf{Dual-VaDE (ours)} & \textbf{ReIm} & \ding{51} & $\mathbf{0.648}${\scriptsize$\,\pm$0.016}$^{\phantom{\ddagger}}$ & $\mathbf{0.610}${\scriptsize$\,\pm$0.018}$^{\phantom{\ddagger}}$ & $\mathbf{0.447}${\scriptsize$\,\pm$0.011}$^{\phantom{\ddagger}}$ & $\mathbf{0.451}${\scriptsize$\,\pm$0.028}$^{\phantom{\ddagger}}$ & $\underline{0.449}${\scriptsize$\,\pm$0.026}$^{\phantom{\ddagger}}$ & $\mathbf{0.562}${\scriptsize$\,\pm$0.024}$^{\phantom{\ddagger}}$ & \textbf{1.08} \\
\midrule
\multicolumn{3}{@{}l}{\textit{Relative change of ours vs.\ VaDE}} & $+39.1\%$ & $+93.7\%$ & $+13.7\%$ & $-46.4\%$ & $-61.9\%$ & $-21.6\%$ & \\
\bottomrule
\end{tabular}
\end{table*}

\subsection{Inference and objective}
\label{sec:dual-domain-architecture}

The two encoders share a single decoder and yield $q_{\phi_s}(\vv z_s\mid\vv x)=\mathcal{N}(\bm\mu_s,\mathrm{diag}\,\bm\sigma_s^2)$ and $q_{\phi_f}(\vv z_f\mid\vv x)=\mathcal{N}(\bm\mu_f,\mathrm{diag}\,\bm\sigma_f^2)$. The push-forward through $\vv A$ gives the fused posterior $q(\vv z^\star\mid\vv x)=\mathcal{N}(\bm\mu^\star,\vv D^\star)$ with $\bm\mu^\star=\beta \vv W_s\bm\mu_s+\alpha \vv W_f\bm\mu_f$ and $\log\bm\sigma^{\star 2}=\vv G_s\log\bm\sigma_s^2+\vv G_f\log\bm\sigma_f^2$, where the learnable weights $\vv G_s,\vv G_f$ sum to one. Training minimizes the dual-domain negative ELBO
\begin{equation}
\begin{aligned}
\mathcal{L}^\star =\;& \mathbb{E}_{q(\vv z^\star\mid \vv x)}\big[\|\vv x - p_\theta(\vv z^\star)\|^2\big] + \lambda_c \mathrm{KL}(\bm\gamma\,\|\,\bm\pi) \\
&+ \lambda_z \textstyle\sum_{k=1}^{K}\gamma_k\,\mathrm{KL}\big(\mathcal{N}(\bm\mu^\star,\vv D^\star)\,\|\,\mathcal{N}(\vv m_k^\star,\bm\Sigma_k^\star)\big).
\end{aligned}
\label{eq:new_loss}
\end{equation}
In contrast to Eq.~\eqref{eq:standard_elbo}, the KL term is evaluated against the structured components of Eq.~\eqref{eq:induced}, so that the spatial and spectral codes are coupled through $\vv C_k$. The mixture parameters are updated by expectation-maximization~\cite{dempster1977em} between gradient steps. The E-step computes responsibilities in the fused space,
\begin{equation}
\gamma_{ik}=\pi_k\,\mathcal{N}(\bm\mu^\star_i\mid\vv m^\star_k,\bm\Sigma^\star_k)\,\big/\textstyle\sum_{j}\pi_j\,\mathcal{N}(\bm\mu^\star_i\mid\vv m^\star_j,\bm\Sigma^\star_j),
\label{eq:estep}
\end{equation}
whereas the M-step operates in the joint space, so that every block of $\tilde{\bm\Sigma}_k$, including $\vv C_k$, is estimated. With $N_k=\sum_i\gamma_{ik}$ and $\bm\zeta_i=[\bm\mu_{s,i};\bm\mu_{f,i}]$,
\begin{equation}
\pi_k=N_k/N,\qquad \tilde{\vv m}_k=\textstyle\frac{1}{N_k}\sum_i\gamma_{ik}\bm\zeta_i,
\label{eq:mstep_mean}
\end{equation}
\begin{equation}
\tilde{\bm\Sigma}_k=\textstyle\frac{1}{N_k}\sum_i\gamma_{ik}(\bm\zeta_i-\tilde{\vv m}_k)(\bm\zeta_i-\tilde{\vv m}_k)^{\top},
\label{eq:mstep_cov}
\end{equation}
and Eq.~\eqref{eq:induced} maps the updated parameters to the decoder input.

\section{Experiments}
\label{sec:experiments}
\noindent\textbf{Data.} We use 253 OASIS-3 brain MR images~\cite{lamontagne2019oasis3} from Alzheimer's patients aged 63--75 to reduce age/diagnosis confounding and capture within-disease anatomical heterogeneity. Images are registered, resampled to $128\times128$, and split into 202 train and 51 test.

\noindent\textbf{Implementation, Baselines \& Metrics.} For a controlled comparison, all models share the same encoder/decoder and training schedule; they differ only in which latent code uses a mixture prior. We train with seeds 42/92/123 and $K\in\{3,5,7\}$. After training, we fit a $K$-component GMM to posterior means on the training set and assign each test image to a prototype. 

We compare \emph{VaDE}~\cite{jiang2017vade} (mixture prior on the spatial code), \emph{Freq-VaDE} (published spectral encoder), and \emph{Dual-VaDE} (our spatial branch plus the dual-domain prior of Sec.~\ref{sec:method} on the same spectral encoder). We test four Fourier inputs: full magnitude (FullFreq)~\cite{frank2020leveraging}, low-frequency magnitude (LowFreq)~\cite{li2021fno,rippel2015spectral}, full complex spectrum with complex convolutions (PolarFreq)~\cite{trabelsi2018complex,gopalakrishnan2024complex}, and real/imaginary channels with real convolutions (ReIm)~\cite{chi2020ffc,stanic2023complex}. Unless noted, we use Dual-VaDE with ReIm. We report test Silhouette~\cite{rousseeuw1987silhouettes} and Davies-Bouldin (DB)~\cite{davies1979cluster} scores (mean $\pm$ SD over seeds) and compare to baselines with a one-sided Welch’s $t$-test~\cite{welch1947}. Inference time is averaged over ten runs on an NVIDIA RTX PRO 6000 Blackwell GPU.

\begin{figure*}[!t]
\centering
\includegraphics[width=.98\textwidth]{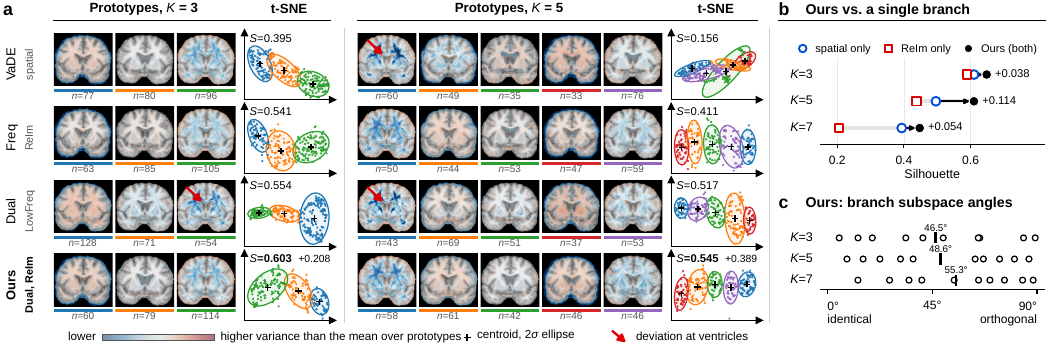}
\caption{(a) Prototypes with relative-variance overlays (red: above mean, blue: below; red arrows: large ventricular deviations) and t-SNE for $K=3$ and $K=5$. Labels match ours, so colors indicate the same prototype across rows. $n$: images per prototype; $S$: latent separation. (b) Ours vs. single-branch variants. (c) Principal angles between our spatial and spectral latent subspaces (dots) and their mean (bars).}
\label{fig:mean-logratio}
\end{figure*}

\noindent\textbf{Prototype separation.}\label{sec:quantitative} In Tab.~\ref{tab:oasis-full}, Dual-VaDE achieves the best Silhouette for all $K$ ($0.648$, $0.610$, $0.447$) and the lowest DB at $K=3$ and $7$ (mean rank $1.08$ vs.\ $4.08$ for the runner-up); at $K=5$ its DB ($0.449$) ties Freq-VaDE+LowFreq. Versus VaDE, it raises Silhouette by $39.1\%$, $93.7\%$, and $13.7\%$ and lowers DB by $46.4\%$, $61.9\%$, and $21.6\%$ at $K=3$, $5$, and $7$, with significant gains ($p<0.05$) in 37/48 comparisons. Fusion effects vary by phase: moving from Freq-VaDE to Dual-VaDE, magnitude-only FullFreq/LowFreq reduce Silhouette at $K=3$ and $5$, while phase-preserving PolarFreq/ReIm improve for all $K$, suggesting magnitude overlaps with the spatial code but phase provides complementary information. LowFreq is strong for $K\leq5$ but fails at $K=7$ (Silhouette $0.058$, DB $2.003$).

\noindent\textbf{Prototypes and latent structure.}\label{sec:mean-logratio} Fig.~\ref{fig:mean-logratio}(a) compares learned prototypes and latent embeddings for our model, two single-domain variants, and the strongest baseline (Dual-VaDE with LowFreq), using one seed per $K$ and Hungarian-matched labels~\cite{kuhn1955hungarian}. The t-SNE~\cite{vandermaaten2008tsne} plots align with Tab.~\ref{tab:oasis-full}: VaDE prototypes overlap ($S=0.156$), while ours are more separated, especially at $K=3$ ($S=0.603$) and $K=5$ ($S=0.545$). At $K=5$ (VaDE) and for both $K$ in Dual-VaDE with LowFreq, some prototypes show large variance deviations near the lateral ventricles and periventricular white matter (red arrows), consistent with enlargement markers~\cite{nestor2008ventricular}. Our deviations are smaller and more uniform, and the prototypes remain sharp.

\noindent\textbf{Efficiency, robustness, and complementarity.}\label{sec:reim-ablation} Fig.~\ref{fig:model} (right) shows that our model is the most accurate and among the fastest spectral models: mean Silhouette $0.568$ vs.\ $0.493$ next best, at $0.277$~ms/image (Freq-VaDE with ReIm: $0.276$~ms), and $20\%$/$22\%$ faster than PolarFreq/LowFreq. It is also the most robust, with the lowest across-seed SD ($0.015$ averaged over $K$ vs.\ $0.020$--$0.039$) and the best worst-case over $K$ ($0.447$). 

Fig.~\ref{fig:mean-logratio}(b) shows that the full model beats the better single branch by $0.038$, $0.114$, and $0.054$ Silhouette at $K=3$, $5$, and $7$ (8/9 seeds). Fig.~\ref{fig:mean-logratio}(c) links this to complementarity: principal angles~\cite{bjorck1973angles} between the top-10-PC subspaces of spatial and spectral latent means average $46.5^\circ$, $48.6^\circ$, and $55.3^\circ$ for $K=3$, $5$, and $7$, with some near-orthogonal directions. Thus, the branches capture distinct factors, making the cross-domain block $\vv C_k$ informative.

\section{Conclusion \& Discussion}
\label{sec:conclusion}
This work introduces Dual-VaDE, an unsupervised deep generative framework that jointly models latent mixture distributions across coupled spatial and Fourier domains.
By explicitly encoding cross-domain covariance and phase-preserving spectral components, the formulation regularizes the latent manifold against high-frequency blur and prevents mode collapse. Model parameters are inferred through an analytically grounded expectation-maximization scheme tailored for the coupled dual-domain objective.
Evaluated on real brain MRI cohorts, Dual-VaDE uncovers distinct neuroanatomical subtypes with sharp cortical and ventricular boundaries while substantially mitigating the random-seed sensitivity common to deep mixture clustering.
These geometric and statistical gains are achieved without prohibitive parameter scaling or inference bottlenecks. Future work will extend this dual-domain mixture formulation to continuous spatiotemporal manifolds, establishing a unified generative prior for mapping longitudinal structural atrophy trajectories.
\section{Acknowledgments \& Ethics}
\label{sec:ethics}
This research was conducted using open-access data resources and received no external funding.
{\small
\bibliographystyle{IEEEbib}
\bibliography{strings,refs}
}

\end{document}